%% file: main.tex
\documentclass[pmlr]{jmlr}

\usepackage{longtable}

 \usepackage{booktabs}
 
\usepackage[load-configurations=version-1]{siunitx} 
\usepackage{import}
\usepackage{multirow}
\usepackage{booktabs}
\usepackage{amsmath}
\usepackage{algorithm}
\usepackage{algorithmic}
\usepackage[inkscapelatex=true]{svg}
\usepackage{comment}
\usepackage{float}
\usepackage[normalem]{ulem}

\usepackage{todonotes}
\makeatletter
\def\set@curr@file#1{\def\@curr@file{#1}} 
\makeatother

\theorembodyfont{\upshape}
\theoremheaderfont{\scshape}
\theorempostheader{:}
\theoremsep{\newline}

\jmlrvolume{182}
\jmlryear{2022}
\jmlrworkshop{Machine Learning for Healthcare}

\title[Leveraging Hierarchy for Curriculum Learning in Automated ICD Coding]{HiCu: Leveraging Hierarchy for Curriculum Learning in Automated ICD Coding}

\author{\Name{Weiming Ren}
        \Email{wren@cs.toronto.edu}
        \AND
        \Name{Ruijing Zeng}
        \Email{jackzeng@cs.toronto.edu}
        \AND
        \Name{Tongzi Wu}
        \Email{tongziwu@cs.toronto.edu}
        \AND
        \Name{Tianshu Zhu}
        \Email{tianshu@cs.toronto.edu}
        \AND
        \Name{Rahul G. Krishnan}
        \Email{rahulgk@cs.toronto.edu}\\ 
        \addr Department of Computer Science,  University of Toronto \& the Vector Institute\\
       Toronto, Ontario, Canada} 

\begin{document}

\maketitle

\begin{abstract}
There are several opportunities for automation in healthcare that can improve clinician throughput. One such example is assistive tools to document diagnosis codes when clinicians write notes. We study the automation of medical code prediction using curriculum learning, which is a training strategy for machine learning models that gradually increases the hardness of the learning tasks from easy to difficult. One of the challenges in curriculum learning is the design of curricula -- i.e., in the sequential design of tasks that gradually increase in difficulty. We propose Hierarchical Curriculum Learning (HiCu), an algorithm that uses graph structure in the space of outputs to design curricula for multi-label classification. We create curricula for multi-label classification models that predict ICD diagnosis and procedure codes from natural language descriptions of patients. By leveraging the hierarchy of ICD codes, which groups diagnosis codes based on various organ systems in the human body, we find that our proposed curricula improve the generalization of neural network-based predictive models across recurrent, convolutional, and transformer-based architectures. Our code is available at \url{https://github.com/wren93/HiCu-ICD}.
\end{abstract}

\import{}{1_introduction}

\import{}{2_related_work}

\import{}{3_methodology}

\import{}{4_icd_coding}

\import{}{5_experiment}

\import{}{6_discussions}


\bibliography{ref}

\import{}{appendix}

\end{document}

%% file: 1_introduction.tex
\section{Introduction}

The prediction of multiple labels from an input occurs in various problems in machine learning. These can include the prediction of multiple clinical pathologies in a chest x-ray \citep{wang2017chestx}, the identification of named entities in biomedical text \citep{dougan2014ncbi}, and, as we study here, the prediction of diagnosis codes from clinical notes \citep{johnson2016mimic}. 
Multi-label classification can be challenging since the accurate prediction of multiple labels requires the ability to effectively learn different decision boundaries (that may depend on different combinations of overlapping features) for each label. Furthermore, when the label space is large, models may focus on accurately predicting commonly observed labels at the expense of rare labels. 

We study the automation of predicting medical diagnosis codes from clinical notes. 
We focus on International Classification of Diseases (ICD) coding, a multi-label classification task. ICD codes are used in healthcare systems to record the diagnoses and procedures during patient stays in medical institutions. After the patient is discharged from the hospital, it is common to assign ICD codes based on the discharge summaries to the patient’s profiles, such as electronic health records (EHR). This task is useful for a variety of purposes such as medical research, epidemiological studies, and billing\footnote{\url{https://en.wikipedia.org/wiki/Clinical\_coder}}. 
Traditionally, this task is often done by professional clinical coders but is  time-consuming and error-prone. Training professional coders is time and labor-intensive. Research such as ours can improve automated ICD coding. One immediate use case is to assist hospital administrators in automating coding from clinical notes. A more long-term use case we envision is that such models may be deployed in auto-complete software that tags the relevant ICD code while the care provider is writing in the original note. Such tools can go a long way toward reducing costs for patients and providers. By reducing clinician time spent on administrative tasks, such automation can help clinicians spend more on patient care and consequently assist in reducing burnout.


Recent work for this problem has focused on developing new neural architectures for this task. However, this is a challenging learning problem -- existing approaches typically treat each label as an independent prediction problem \citep{tsoumakas2007multi} and can underperform in the prediction of less frequently observed diagnosis codes. A critical limitation to many deep learning approaches to this problem has been that they ignore the structure inherent among the different labels. This structure may already exist among the labels or, optionally, can be extracted by querying domain experts. ICD codes, for example, are organized hierarchically. Our focus in this work is to leverage the structure among labels to simplify the learning problem.

When tackling challenging learning problems, an effective strategy can be to leverage curriculum learning (CL) \citep{bengio2009curriculum}. Instead of solving a single learning problem, CL posits that we may improve a model's generalization by training it iteratively using a curriculum -- first on simple problems and then on more challenging ones. Using CL requires knowing what constitutes good curricula, a complex problem in and of itself. 

\begin{figure}[]
  \centering
  \includegraphics[width=1.0\textwidth]{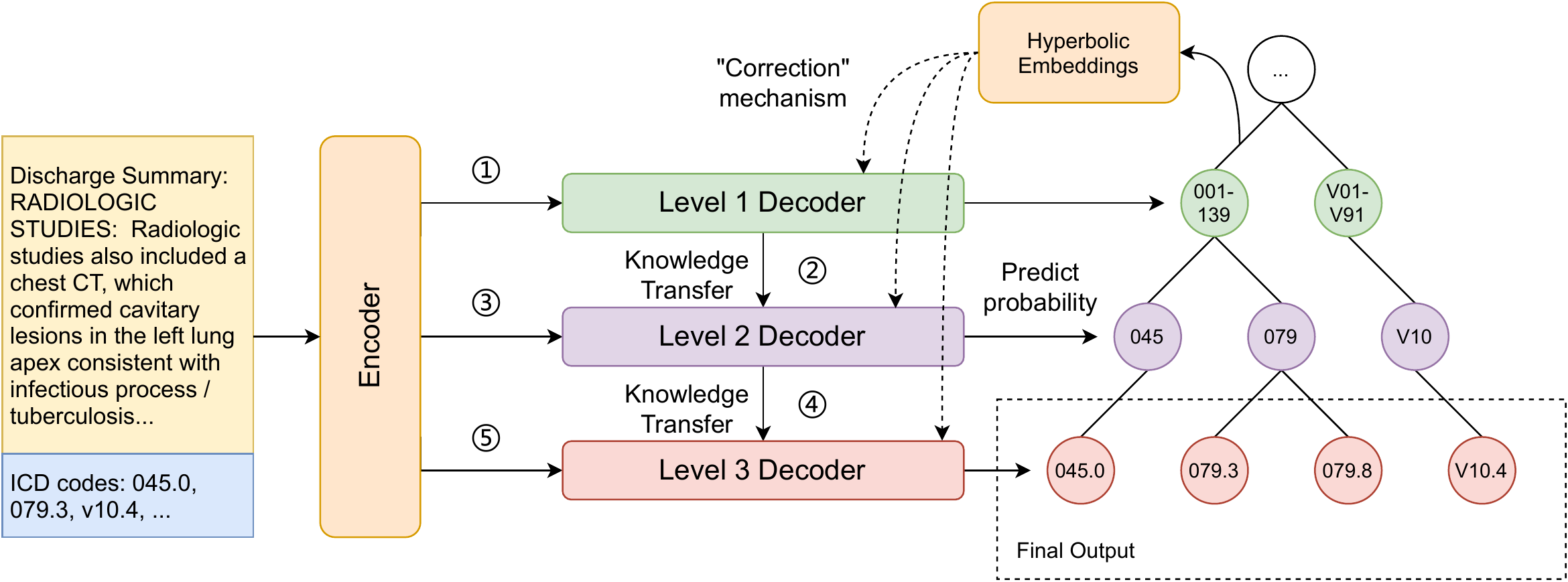}
  \caption{\small \textbf{Hierarchical Curriculum Learning (HiCu):} Illustration of the overall encoder-decoder architecture and the hierarchical curriculum learning algorithm of our ICD coding model. The numbers indicate the sequential execution order of our training algorithm: the model is first trained on labels at the first level of the label tree using level one decoder, and then proceeds to level two using the knowledge transfer mechanism. This process is repeated until the model reaches the final level in the label tree. During training at each level, the hyperbolic embeddings of the ICD codes are used to guide the attention computation in the decoder.}
  \label{fig:overall}
\end{figure}

In this work, we tackle the challenges of multi-label classification using curriculum learning and experiment on the MIMIC-III v1.4 dataset with ICD-9 codes, which is the primary benchmark dataset for ICD coding classification. Our first insight is that the decision boundaries for different ICD codes are not independent. Instead, ICD codes are organized in a tree structure which defines a notion of similarity, i.e. dissimilar labels will have different ancestors in the tree and vice versa. In addition, the specificity of codes increases the further down we go into each sub-tree. Next, we note that ICD codes in a population follow a power-law distribution -- this means the labels tend to be imbalanced with a few handfuls of commonly used codes and a long tail of rare codes. Our second insight is that the explicit incorporation of techniques to handle label imbalance is essential to ensure parity of performance of predictive models on both rare and frequent labels. The main contributions of this paper are as follows:
\begin{enumerate}
    \item We propose a novel \textbf{Hi}erarchical \textbf{Cu}rriculum learning algorithm \textbf{HiCu} (see Figure~\ref{fig:overall}) to learn multi-label classifiers when labels exhibit a hierarchical structure. HiCu creates a curriculum for learning based on a depth-wise decomposition of the label graph and a hyperbolic-embedding-based knowledge transfer mechanism to warm start representations from one step of curriculum learning to the next. We incorporate an asymmetric loss function into our model to deal with the highly imbalanced label distribution to help balance the performance on both frequent and rare labels. 
    \item We thoroughly evaluate the algorithm with ablation studies to quantify the impact of each of our contributions and show that HiCu dramatically improves multi-label predictive performance on ICD code prediction across recurrent, convolutional, and transformer-based neural architectures.
\end{enumerate}


\subsection*{Generalizable Insights about Machine Learning in the Context of Healthcare}


Our work provides a method to leverage curriculum learning for multi-label classification tasks where the label space is structured. Specifically, we consider label spaces that are structured as a hierarchy, a directed graph where the specificity of the label increases with depth in the graph. A practical ramification of using this structure is that the resulting method improves predictive performance on less frequently occurring (and consequently harder to detect) labels. We instantiate our algorithm for automating the prediction of ICD codes helping accelerate the development of automated solutions to identify diagnosis codes from clinical notes such as discharge summaries. For a fair comparison to previous work, we run thorough experiments with standard benchmarks using ICD-9 codes and a variety of different neural architectures. Since the hierarchy of the ICD-10 codes is similar to the ICD-9 hierarchy, we anticipate our algorithm, HiCu, to continue to perform well in that domain.

The idea behind our work may be extended to other multi-label classification problems with structure among the labels. One such example is OncoTree, a tree-based ontology of 868 tumor types across 32 organ sites. An interesting new application of HiCu that would be fertile ground for future work would be to leverage hierarchical ontologies to regularize representations in predictive tasks. For example, with OncoTree, one could help improve predictive models of cancer severity from histopathological tissue samples while leveraging the hierarchy to predict tumor type as a form of regularization.


%% file: 2_related_work.tex
\section{Related Work}
\textbf{Curriculum Learning:} The design of algorithms has often been inspired by human learning. 
Human education is organized sequentially, starting with simple concepts, which are composed to form more sophisticated concepts \citep{wang2021survey}.
In a similar vein, \citet{bengio2009curriculum} proposed curriculum learning (CL) for learning models. 
Typically curriculum learning \citep{elman1993learning} breaks up the training data into subsets and learns the model first on the easier sets of data and then on the harder ones. CL has been used in computer vision (CV) \citep{guo2018curriculumnet, jiang2014easy}, natural language processing (NLP) \citep{platanios2019competence, tay2019simple}, and various reinforcement learning (RL) tasks \citep{florensa2017reverse, narvekar2017autonomous, ren2018self}. In the context of healthcare, \citep{el2020student} designs a student-teacher framework where a model adaptively creates curricula by sub-selecting the next set of inputs for a student network to be trained on. CL often results in a faster convergence rate during training and improved generalization.
\citet{platanios2019competence} employed CL to aid a neural machine translation model and cut training time by up to 70\%. 
\citet{florensa2017reverse} used CL to help RL agents answer difficult goal-oriented problems that they couldn't handle without the help of curriculum. 
Our work differs from existing work in a few ways. We define curricula, not over the set of inputs that a model is trained with at each step but rather the span of labels the model has to predict. Next, we leverage privileged-information \cite{vapnik2015learning} (information available at training time but not at test time) in the design of curricula resulting in improvements in the model generalization that we examine both quantitatively and qualitatively.

\textbf{Hierarchical Knowledge \& Hyperbolic Embedding:} Some recent approaches have successfully demonstrated the benefits of leveraging hierarchical knowledge for training machine learning models. For example, \citet{zhang2019ernie} utilizes both large-scale textual corpora and knowledge graphs (KGs) to train an Enhanced Language Representation with Informative Entities (ERNIE), which can take full advantage of lexical, syntactic, and knowledge information simultaneously. Hyperbolic embeddings, a special case of hierarchical knowledge, have captured the attention of the machine learning community. The motivation is to embed structured, discrete objects such as knowledge graphs into a continuous representation that can be used with modern machine learning methods. Hyperbolic embeddings can preserve graph distances and complex relationships in very few dimensions, particularly for hierarchical graphs \citep{nickel2017poincare, chamberlain2017neural}.

For example, given a tree, Hyperbolic Embeddings for Entities (HyperE) \citep{sala2018representation} used combinatorial construction that embeds the tree in hyperbolic space with arbitrarily low distortion without using optimization. HyperE offers excellent quality with few dimensions when embedding hierarchical data structures like synonyms or type hierarchies. Hyperbolic Attention Networks, proposed by \citet{gulcehre2018hyperbolic}, imposed hyperbolic geometry on the activations of neural networks. This allows the model to exploit hyperbolic geometry to reason about embeddings produced by deep networks. Hyperbolic and Co-graph Representation for Automatic ICD Coding (HyperCore) \citep{cao2020hypercore} utilized the idea of hyperbolic embeddings to improve the performance of automated ICD coding. Their proposed model outperforms previous state-of-the-art methods by utilizing hyperbolic embeddings to capture the code hierarchy. 

\textbf{Automated ICD coding:} The automated ICD coding of clinical notes is an established studied research topic \citep{perotte2014diagnosis, koopman2015automatic, karimi2017automatic, shi2017towards, mullenbach2018explainable, baumel2018multi, li2018automated, xie2019ehr, huang2019empirical, chen2019multi, li2020icd, vu2020label}. \citet{larkey1996combining} used an ensemble model combining KNN, relevance feedback, and Bayesian independence classifiers to predict ICD-9 codes based on discharge summaries.
\citet{de1998hierarchical} utilized the cosine similarity between the medical discharge summary and the ICD code description to build the classifier which assigns codes with the highest similarities to the summary. \citet{perotte2014diagnosis} proposed a supervised machine learning model that used SVM to perform automatic ICD coding on the MIMIC-II dataset. 

Recently, this problem has been explored using tools from deep learning. \citet{shi2017towards} used character-level long short-term memory (LSTMs) to learn representations of subsections from discharge summaries and the code description, followed by an attention mechanism to address the mismatch between the subsections and corresponding codes. 
\citet{wang2018joint} proposed a joint embedding model, in which the labels and words are embedded into the same vector space, and the cosine similarity between them is used to predict the labels. \citep{baumel2018multi} proposed a hierarchical gated recurrent unit (GRU) network for assigning multiple ICD codes to discharge summaries from the MIMIC II and III datasets. For convolutional neural network (CNN) based methods, \citet{mullenbach2018explainable} used a one-layer CNN in conjunction with a structured attention mechanism to predict ICD-9 codes from multiple-labeled clinical notes. \citet{xie2019ehr} improved upon the convolutional attention model \citep{mullenbach2018explainable} by using densely connected CNN and multi-scale feature attention. The convolutional attention model \citep{mullenbach2018explainable} was further improved again by \citet{li2020icd}, who proposed a Multi-Filter Residual CNN (MultiResCNN) for ICD coding. 
\citet{vu2020label} additionally proposed a hierarchical joint learning mechanism extending their label attention model using the hierarchical relationships among the codes. To handle the issue of label imbalance, \citet{kim2021read} introduced the Read, Attend, and Code (RAC) model for learning ICD code assignment mappings. 
Our work explores an orthogonal direction to the status quo. Rather than devising a new neural architecture for this problem, we study the improvements that may be realized through curriculum learning.

%% file: 3_methodology.tex
\section{Methodology}
Automatic ICD coding can be regarded as a multi-label classification problem \citep{mccallum1999multi}. Formally, assume we have a dataset $\{\mathbf{x}, \mathbf{y}\}_{i=1}^D$, where $\mathbf{x} \in \mathbb{R}^N$ denotes an input discharge summary represented by word indices and $\mathbf{y} \in \{0, 1\}^{|\mathbf{C}|}$ denotes the set of binary indicators of all the ICD codes;
our goal is to train $|\mathbf{C}|$ binary classifiers, each predicting the probability of assigning an ICD code to the input discharge summary.
Our model, like others, is an encoder-decoder architecture \citep{mullenbach2018explainable, li2020icd}. Given an input discharge summary, the encoder creates feature representations for each word in the input. The decoder then maps these representations onto ICD codes.

\subsection{Encoder}
\label{sec:encoder}
Let $\mathbf{x} \in \mathbb{R}^N$ be the input word sequence represented by word indices, the encoder takes in this input and outputs a word representation matrix $\mathbf{H} \in \mathbb{R}^{N\times d_f}$, where each row $H_i \in \mathbb{R}^{d_f}$ represents a feature vector of the word $i$. The goal of any encoder is to extract semantically meaningful representations from the input word sequences. This can be any text feature extractor such as CNN-based models \citep{mullenbach2018explainable, li2020icd}, RNN-based models \citep{vu2020label} and transformer-based pre-trained language models \citep{pascual2021towards}. In this work, we study our method using three existing architectures, namely Bi-LSTM \citep{vu2020label}, MultiResCNN \citep{li2020icd} and RAC reader \citep{kim2021read}, to directly compare our results with prior works. A detailed introduction of these encoder architectures can be found in Section~\ref{sec:encoder_arch}.

\subsection{Decoder}
\label{sec:decoder}
Following \citet{mullenbach2018explainable}, our decoder contains a per-label attention network for aggregating output features from the encoder and a classification head for final ICD code prediction. Applying a per-label attention network to the encoder output lets each ICD code attend to different parts of the token sequence and automatically find the most relevant tokens to perform prediction. The per-label attention layer is:
\begin{equation}\label{equ:attention}
    \mathbf{A} = softmax(\mathbf{H}\mathbf{Q}) \in \mathbb{R}^{N\times |\mathbf{C}|},
\end{equation}
\begin{equation}
    \mathbf{V} = \mathbf{A}^\top \mathbf{H} \in \mathbb{R}^{|\mathbf{C}| \times d_f},
\end{equation}
where $\mathbf{Q} \in \mathbb{R}^{d_f \times |\mathbf{C}|}$ is the query matrix for each ICD code. Column $i$ in $\mathbf{Q}$ represents the query vector for ICD code $i$. $\mathbf{A}$ represents the attention score of each token with respect to each ICD code and $\mathbf{V}$ is the output matrix computed from $\mathbf{A}$ and the encoder output $\mathbf{H}$. Each row in $\mathbf{V}$ represents the weighted sum of the token embeddings for an ICD code based on the attention scores. These aggregated embeddings are then fed into a classification head to perform a final linear transformation and sum pooling \citep{li2020icd} to get the raw output $\tilde{y} \in \mathbb{R}^{|\mathbf{C}|}$:
\begin{equation}
    \mathbf{Z} = \mathbf{V}\mathbf{W} \in \mathbb{R}^{|\mathbf{C}| \times |\mathbf{C}|},
\end{equation}
\begin{equation}
    \tilde{y} = sum\_pool(\mathbf{Z}) + b ,\ \ \tilde{y}_i = \sum_{j=1}^{|\mathbf{C}|}\mathbf{Z}_{ij} + b_i
\end{equation}
where $\mathbf{W} \in \mathbb{R}^{d_f\times|\mathbf{C}|}$ is the weight matrix of the final linear layer and $b$ is the bias term. The output logits are computed via the Sigmoid function: $\hat{y} = \sigma(\tilde{y})$.

\subsection{HiCu: Hierarchical Curriculum Learning}
\label{sec:hcl}
We now present our hierarchical curriculum learning algorithm for automated ICD coding. We first describe a generic learning algorithm for multi-label classification. In Section~\ref{sec:icd_coding}, we illustrate why ICD coding can be benefited from hierarchical curriculum learning and how to apply our algorithm to automate ICD coding from discharge summaries. Figure~\ref{fig:overall} illustrates the overall procedure of our hierarchical curriculum learning algorithm. Given the task of training $|\mathbf{C}|$ binary classifiers over a label set $\mathbf{C}$ based on input $\mathbf{x}$, our algorithm solves the problem using the following steps.
\begin{itemize}
    \item We construct a label tree from the label set $\mathbf{C}$, where each label $c \in \mathbf{C}$ is located at the last level (leaf node) in the label tree.
    \item We then generate a hierarchical training curriculum and sequentially train our model to predict the labels at each level of the label structure. During the training process, the weights in the per-label attention network at each level are initialized using a knowledge transfer process and guided by hyperbolic embeddings.
    \item The model outputs predictions of the labels at the last level of the tree structure, which are the labels in the target label set $\mathbf{C}$.
\end{itemize}

\subsubsection{Generating Curricula from Label Trees}
The key idea that we exploit is that structure in the output space of labels is a natural lever that practitioners can leverage to create ordered tasks from easy to difficult. Our conjecture is the algorithm results in helping models find better local minima by iteratively solving tasks and building up to the final one rather than solving the final task directly.
This learning curriculum is generated from a label tree constructed from the target label set $\mathbf{C}$. For label $c_i$ in the label set, we define a path $P_i=\langle V_i, E_i\rangle$ that starts from the root node and ends at label $c_i$:
\begin{align}
    V_i=\{root, c^{(1)}_i, c^{(2)}_i, ..., c^{(K_i)}_i\}, c^{(K_i)}_i=c_i, \\
    E_i = \{(root, c^{(1)}_i), (c^{(1)}_i, c^{(2)}_i),...,(c^{(K_i-1)}_i, c^{(K_i)}_i) \},
\end{align}
where $V$ and $E$ denote the set of the vertices (nodes) and edges along this path, respectively. The label tree can thus be defined as $T = \langle V, E\rangle$, where $V = \bigcup\limits_i V_i$ and $E = \bigcup\limits_i E_i$.

The length of the paths for different labels may not always be the same -- the target labels may be located at different levels after the label tree is constructed, making it difficult to design learning curricula based on the label tree. To solve this problem, for each label $c_i$ in the label set, we introduce an augmented path $\hat{P}_i=\langle \hat{V}_i, \hat{E}_i \rangle$ whose length is always equal to the maximum length among all root-to-label paths in the label tree $T$:
\begin{equation}
    \hat{V}_i = \{root, c_i^{(1)}, ..., c_i^{(K_i)}, c_i^{(K_i+1)}, ..., c_i^{(K_{max})}\},c_i^{(K_i)}=c_i^{(K_i+1)}=...=c_i^{(K_{max})}=c_i,
\end{equation}
\begin{equation}
    \hat{E}_i = \{(root, c^{(1)}_i), (c^{(1)}_i, c^{(2)}_i),..., (c^{(K_i)}_i, c^{(K_i+1)}_i), ..., (c^{(K_{max}-1)}_i, c^{(K_{max})}_i) \},
\end{equation}
where $K_{max}$ is the maximum possible length for all root-to-label paths in the label tree $T$. We then define the augmented label tree $\hat{T}=\langle \hat{V}, \hat{E}\rangle$, where $\hat{V} = \bigcup\limits_i \hat{V}_i$ and $\hat{E} = \bigcup\limits_i \hat{E}_i$. This means that target nodes in the middle of the tree are repeated in the augmented tree in every subsequent layer. In this way, all the target labels will be located at level $K_{max}$ under the root node of the label tree and the multi-label classification problem over the target label set $\mathbf{C}$ can, at the final level, corresponds to $|\mathbf{C}|$ binary classification problems.


\subsubsection{Knowledge Transfer with Hyperbolic Correction}
Given an augmented label tree $\hat{T}$ with $K$ levels under the root node, our curriculum learning algorithm is formulated as a $K$-round training over the entire tree structure. For round $k$ in the training process, our training target is to predict the labels at the $k^{th}$ level of the augmented label tree. 
Before moving onto the $(k+1)^{th}$ level, we need to map from the parameters of the current model onto a new one that predicts over the labels at the $(k+1)^{th}$ level. We define the following knowledge transfer function that maps from the parameters of the model at the $k^{th}$ level onto one that predicts labels at the $(k+1)^{th}$ level.

\begin{itemize}
    \item For the encoder, parameters for round $k+1$ are initialized to be the same as the encoder parameters after round $k$. In other words, for each round in the training process, the encoder is fine-tuned based on the pre-trained weights of the previous round. This transfers knowledge from the previous round and provides a sound initialization of the model parameters when the training proceeds to predict the actual target labels.
    \item For the decoder, we initialize the query matrix $\mathbf{Q}^{(k+1)}$ in the per-label attention network as described in Equation~\ref{equ:attention} for round $k+1$ using the following equation:
    \begin{equation}
    \label{equ:q_init}
        q^{(k+1)}_{c_i} = q^{(k)}_{c_j}, c_i \in Children(c_j),
    \end{equation}
    where $q^{(k)}_c$ is the query vector ($c^{th}$ column vector in $\mathbf{Q}^{(k)}$) for label $c$ at level $k$. $Children(c)$ represents all the child nodes of the label $c$ in the augmented label tree. The query vectors for the first level are randomly initialized. This process ensures the query vector for similar labels (sibling labels under the same parent label in the augmented label tree) are also similar.
\end{itemize}

It should be noted that a potential problem for training the model using the augmented label tree is that when there are too few splits for a level in the original label tree, padding nodes in the augmented label tree will lead to a large amount of redundant intermediate nodes. As training on these intermediate levels for a label that has appeared before is identical to continuing to fine-tune the binary classifier for that label, this may lead to unnecessary computational overhead and possible overfitting of these labels. To mitigate this issue, we can reduce the number of epochs we train for that level.

\textbf{Hyperbolic Correction: }
The knowledge transfer process in our curriculum learning algorithm implicitly encodes the top-to-bottom hierarchical architecture of the label tree into the query vectors as described in Equation~\ref{equ:attention}. However, initializing the query vectors of all the sibling labels to be the same is not ideal: these labels are similar but not identical. For example, the ICD subtree for type II diabetes contains diverse subcategories based on the complications of diabetes (such as kidney disease and macular edema). 
We would like a way of incorporating bottom-up knowledge about how similar/dissimilar each new child is based on each leaf's subtree.

To solve this problem, we turn to hyperbolic embeddings \citep{nickel2017poincare}, which are representations that are sensitive to structure inherent in a graph.
At the beginning of the training, we pre-train hyperbolic embeddings for all labels in the label tree $T$ using the training method introduced in \cite{nickel2017poincare}. The created hyperbolic embeddings can capture the global structure of the entire ICD tree. After the knowledge transfer mechanism initializes the query vectors, the hyperbolic embeddings are used to build the query vectors at every forward propagation step. We present two methods to generate query vectors:
\begin{itemize}
    \item [1.] \textbf{Addition (HiCuA)} We use a fully connected layer to transform the hyperbolic embedding of a label to a vector with the same dimension as the query vector. This vector is then added to the original query vector at each training step to form the final query vector $\hat{q}_{c_i}$. The final query vector can be described as:
    \begin{equation}
        \hat{q}_{c_i}=q_{c_i} + fc(e_{c_i}),
    \end{equation}
    where $q_{c_i}$ is the original query vector described in Equation~\ref{equ:q_init}, $e_{c_i}$ is the hyperbolic embedding of code $c_i$ and $fc(\cdot)$ is the fully connected layer.
    \item [2.] \textbf{Concatenation (HiCuC)} We concatenate the hyperbolic embedding of a label with the original query vector, and use a fully connected layer to transform the concatenated vector to a new query vector to perform per-label attention. This operation can be described as:
    \begin{equation}
        \hat{q}_{c_i} = fc(q_{c_i} \oplus e_{c_i}),
    \end{equation}
    where $\oplus$ denotes the concatenation operation.
\end{itemize}

The idea behind the correction mechanism is that for each label in the label tree, the original query vector initialized by the knowledge transfer process is served as a ``base'' query vector. However, this query vector ignores the difference between sibling labels, and hyperbolic embedding is used to correct the query vector by taking into account the position of this label in the label tree. The overall learning procedure of our hierarchical curriculum learning algorithm can be found in Algorithm~\ref{alg:hcl}. In Section~\ref{sec:ablation}, we show that both the knowledge transfer initialization and the hyperbolic embedding correction contribute to our model's performance improvement.



%% file: 4_icd_coding.tex
\section{Automated ICD Coding}
\label{sec:icd_coding}
The ICD-9 system classifies diseases in a coarse-to-fine manner. For example, ``Diseases Of The Skin And Subcutaneous Tissue'' (680-709) contains three subcategories:  ``Infections Of Skin And Subcutaneous Tissue'' (680-686), ``Other Inflammatory Conditions Of Skin And Subcutaneous Tissue'' (690-698), and ``Other Diseases Of Skin And Subcutaneous Tissue'' (700-709). These subcategories can be further divided into categories such as ``Carbuncle and furuncle'' (680), ``Cellulitis and abscess of finger and toe'' (681), and ``Other cellulitis and abscess'' (682), etc. As in Figure~\ref{fig:icd_taxonomy}, this disease taxonomy can be naturally represented by a tree structure, making our method a natural fit for automated ICD coding.

\subsection{ICD Code Label Tree}
\begin{figure}[]
  \centering
  \includegraphics[width=1.0\textwidth]{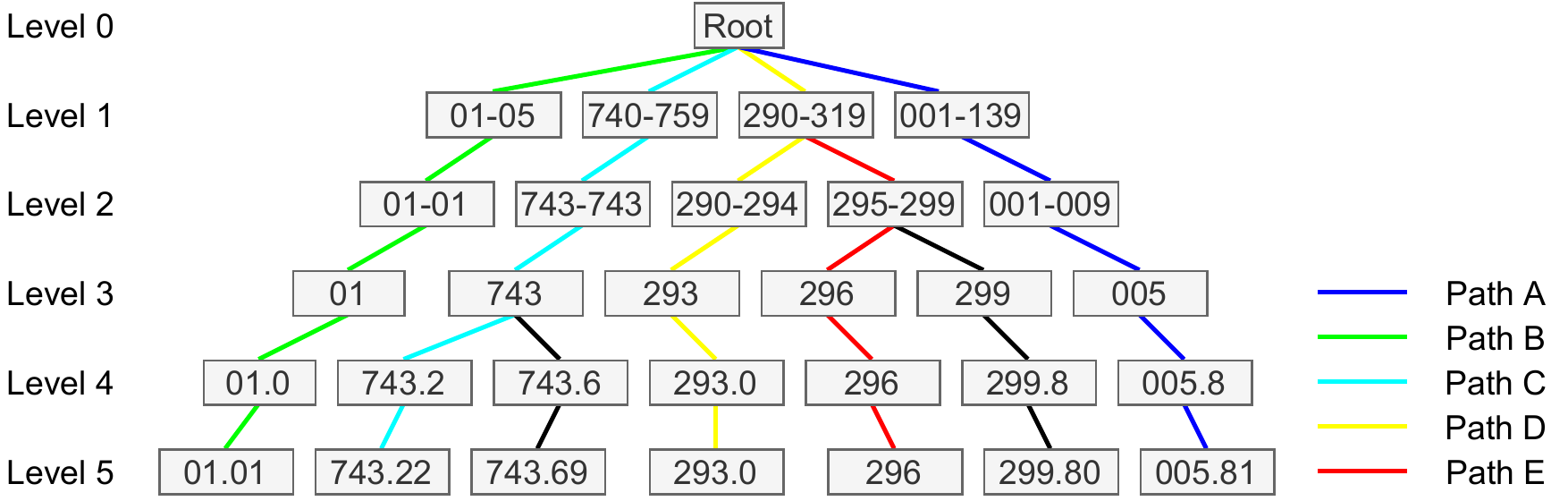}
  \caption{\small A fraction of our ICD code label tree for hierarchical curriculum learning.}
  \label{fig:code_tree}
\end{figure}

Based on the structure of the ICD-9 code system, we design a five-level ICD code label tree ($\textrm{root level}=0$) and fit both diagnosis and procedure codes into this tree. We regard each label in the dataset as a leaf node and construct a path from the leaf node to the root node. As shown by path A in Figure~\ref{fig:code_tree}, The first two levels in the code tree contain ICD code ranges representing the general disease classifications. The third level includes all integer codes, where the diagnosis code is a three-digit number and the procedure code is a two-digit number. The fourth and fifth levels contain ICD codes with one decimal place and two decimal places, indicating finer-grained disease classifications. For some diagnosis codes (e.g., subcodes under 740-759) and all procedure codes, these codes do not contain the second code ranges (level two in the code tree) in the original ICD code system. To align with other codes, we manually create an intermediate level that is also represented using a code range, but the start and end values of this code range are the same, as shown by paths B and C in Figure~\ref{fig:code_tree}. Furthermore, for some labels in the dataset that are integer codes or codes with only one decimal place, we copy the same code for the fourth and fifth levels to generate a full code tree, as shown by paths D and E in Figure~\ref{fig:code_tree}.

\subsection{Learning Objective}
\label{sec:asl}
Prior work \citep{mccallum1999multi, mullenbach2018explainable} has treated the prediction of ICD codes as a multi-label classification problem. For each ICD code, the learning objective is to minimize a binary cross-entropy loss. The overall loss function can be expressed as:
$\mathcal{L} = -\sum_{i=1}^{|\mathbf{C}|}{y_i \log(\hat{y}_i) + (1-y_i)\log(1-\hat{y}_i)},$
where $\hat{y}_i$ is the $i^{th}$ output logit described in Section~\ref{sec:decoder} and $y_i$ is the corresponding ground truth.

While the binary cross-entropy loss is widely used, it may not be a good choice for multi-label classification \citep{DistributionBalancedLoss, oksuz2020imbalance, lin2017focal} when the label space is very large and a very small number of labels are presented for any given example. 
\citet{ben2020asymmetric} proposed the Asymmetric Loss (ASL) that downweights easy negative samples so the model can focus on identifying less frequently occurring positive labels.
The asymmetric loss consists of two mechanisms: asymmetric focusing and probability shifting. It can be expressed as below:
\begin{equation}
\label{equ:asl} 
\mathcal{L} = -\sum_i{y_i (1-\hat{y}_i)^{\gamma_{+}} \log(\hat{y}_i) + (1-y_i) {p_m}_i^{\gamma_{-}} \log(1-{p_m}_i)},
\end{equation}
where $\gamma_{+}$ and $\gamma_{-}$ are focusing parameters, and ${p_m}_i$ is the shifted probability. Note that $\gamma_{+}$ is set to be smaller than $\gamma_{-}$ to emphasize the contribution of rare positive ones. The asymmetric focusing helps the network to achieve a balanced loss generated by the positive and negative samples and contributes to the more meaningful training process. The shifted probability is defined as:
\begin{equation}
{p_m}_i = \max{(\hat{y}_i-m, 0)},
\end{equation}
where the probability margin $m \geq 0$ is a tunable hyperparameter that functions as a threshold to ignore mislabeled negative samples.

For ICD coding, the distribution of ICD codes (the labels) is often very skewed. In practice, patients often exhibit a handful of common codes, but the average ratio between positive and negative samples in each document in our dataset is $15.9:8906.1=0.0018$, which is very imbalanced. Consequently, to focus learning more on positive labels and consequently, we learn with the ASL loss described in Equation \ref{equ:asl}.

%% file: 5_experiment.tex
\section{Experiments} 

\subsection{Dataset}
We used the third version of the \textbf{M}edical \textbf{I}nformation \textbf{M}art for \textbf{I}ntensive \textbf{C}are (MIMIC-III) dataset \citep{johnson2016mimic} to test our models in this study. 

\paragraph{MIMIC-III}
We focus on discharge summaries, which summarize all information during each patient's stay. Every stay was manually labeled by coders using one or more ICD-9 codes, denoting the performed diagnoses and procedures. There are 52,722 discharge summaries and 8,929 unique codes in this dataset. Following the previous work, the experiments were conducted separately by using the full set of codes and using the most frequent 50 codes. For the full-code experiment, the data was separated using patient IDs, ensuring that each patient appeared in only one of the training, validation, or testing sets. There are 47,719 training discharge summaries, 1,631 validation summaries, and 3,372 testing summaries. For the top-50 code experiment, the subset of 11,317 discharge summaries was generated, consisting of 8,067 discharge summaries for training, 1,574 for validation, and 1,730 for testing.

\paragraph{Preprocessing}
Following \citep{li2020icd, mullenbach2018explainable}, the summary text was first preprocessed with only alphabetic characters left and was then transformed into lowercase. 
Then, we pre-train word embeddings using the Word2Vec CBOW method \citep{mikolov2013distributed} on all discharge summaries from the MIMIC-III dataset. Detailed hyperparameters of the Word2Vec embeddings for different encoder architecture can be found in Appendix~\ref{sec:implementation}. The code to reproduce our results is ready for public release.

\subsection{Evaluation Metrics}
We report a variety of different evaluation metrics in our experiments. We focused on the micro-averaged and macro-averaged AUC (area under the ROC curve) and F1 scores. Micro-averaged AUC or F1 are computed by regarding every pair of summary text and codes as a separate prediction, while macro-averaged AUC or F1 are computed by averaging the metrics for each label. The macro-averaged AUC pays more attention to the prediction of rare labels. We also utilized precision at K metrics (P@K) for evaluation, which presented the proportion of accurately predicted labels in the top-K predicted labels. As mentioned in \cite{mullenbach2018explainable}, this was motivated by the reality that users review only a fixed number of ICD codes. In the full-code experiment, we applied the P@5, P@8, and P@15 metrics, while in the 50-code experiment, we employed the P@5 metric. We refer the reader to the supplementary material for details on the implementation.

\subsection{Encoder Architectures}
\label{sec:encoder_arch}
As discussed in Section~\ref{sec:encoder}, we utilize and replicate three encoder architectures to test the effectiveness of HiCu for automated ICD coding:
\paragraph{Bi-LSTM, based on LAAT \citep{vu2020label}}
The Bidirectional Long-Short Term Memory (Bi-LSTM) encoder contains a Word2Vec embedding layer and a Bi-LSTM feature extraction layer. Similar to MultiResCNN, the embedding layer first generates a word embedding for each input token. The Bi-LSTM layer then extracts contextual representations $\overrightarrow{\mathbf{H}}$ and $\overleftarrow{\mathbf{H}}$, where $\overrightarrow{\mathbf{H}}$ is generated by the forward LSTM layer and $\overleftarrow{\mathbf{H}}$ by the backward LSTM layer. The output text representation matrix $\mathbf{H} \in \mathbb{R}^{N\times2u}$ is the concatenation of $\overrightarrow{\mathbf{H}}$ and $\overleftarrow{\mathbf{H}}$, where $u$ is the dimension of the forward/backward LSTM hidden states.

\paragraph{MultiResCNN \citep{li2020icd}}
The Multi-Filter Residual Convolutional Neural Network (MultiResCNN) model is designed based on TextCNN \citep{kim-2014-convolutional} and ResNet \citep{he2016deep}. Given the input word sequence $\mathbf{t}$, it first uses a pre-trained Word2Vec embedding layer to convert the input word indices into a word embedding matrix $\mathbf{E} \in \mathbb{R}^{N\times d_e}$, where $d_e$ is the dimension of the word embeddings. 
The multi-filter residual CNN layer contains $m$ convolutional filters with different kernel sizes. On top of each filter, there is also a residual convolutional layer to enlarge the receptive field of the network. The output $\mathbf{H}_r\in \mathbb{R}^{N\times d_r}$ from each residual convolutional layer are concatenated to form the final output text representation matrix $\mathbf{H} \in \mathbb{R}^{N\times d_f}$, where $d_f = m \times d_r$.

\paragraph{RAC reader \citep{kim2021read}}
The RAC reader is composed of the Convolved Embedding Module and the Self-Attention Module. The Convolved Embedding Module uses a pre-trained Word2Vec embedding layer followed by two CNN layers to representations of text. The output from the Convolved Embedding Module is then sent into the Self-Attention Module, which is a stack of four transformer blocks similar to \citet{vaswani2017attention}. This module extracts word semantics and generates contextual representations. Each transformer block contains one attention head, but unlike the standard transformer architecture, the model is designed to be permutation equivariant\footnote{The output ICD code assignment will not be changed regardless of the order of the input word sequence.}.

\subsection{Baselines}
Other than the models we implemented in Section \ref{sec:encoder_arch}, we also compare our proposed approach against several state-of-the-art baselines for this task.

\paragraph{CAML} 
\textbf{C}onvolutional \textbf{A}ttention network for \textbf{M}ulti-\textbf{L}abel classification model  \citep{mullenbach2018explainable} consists of one convolutional layer and one per-label attention layer, allowing the model to generate label-related features for the multi-label classification. 
 
\paragraph{DR-CAML} 
\textbf{D}escription \textbf{R}egularized \textbf{CAML} is an extension of CAML which utilizes the text description of each code to regularize the model.

\paragraph{MSATT-KG}
\textbf{M}ulti-\textbf{S}cale Feature \textbf{Att}ention and Structured \textbf{K}nowledge \textbf{G}raph Propagation \citep{xie2019ehr} composes a dense convolutional layer and a multi-scale feature attention layer. The hierarchical information among labels is captured during prediction using a graph convolutional network \citep{kipf2016semi}. 

\paragraph{HyperCore}
The \textbf{Hyper}bolic and \textbf{Co}-graph \textbf{Re}presentation method was proposed by \citet{cao2020hypercore}. It contains a hyperbolic embedding layer to leverage the code hierarchy and a graph convolutional network to utilize the code co-occurrence.


\paragraph{JointLAAT}
\textbf{Joint} \textbf{La}bel \textbf{At}tention model is an extension of LAAT \citep{vu2020label} with a hierarchical joint learning mechanism, which was proposed to tackle the problem of imbalanced labels. 




\begin{table}[]
\footnotesize
\caption{\footnotesize 
\textbf{MIMIC-III Full Code Results (in \%).} The results we obtained are shown in means ± standard deviations from 10 random runs. 
The first block in the table shows the baseline results reported in the papers. The following three blocks specify three groups of experiments, each showing the baseline we re-implemented (with *) and the corresponding models we further evaluated with our HiCu learning algorithm. The bold values indicate the best results we obtained from all experiments, whereas the underlined values indicates the best results among each experiment group. 
}
\label{tab:fullcode}
\centering
\begin{tabular}{l|cc|cc|ccc} 
\toprule
\multirow{2}{*}{Model}                  & \multicolumn{2}{c|}{AUC}              & \multicolumn{2}{c|}{F1}                       & \multicolumn{3}{c}{Precision@K}                                        \\ 
\cline{2-8}
                                        & Macro                 & Micro         & Macro                 & Micro                 & P@5                   & P@8                   & P@15                   \\ 
\midrule
CAML                                    & 89.5                  & 98.6          & 8.8                   & 53.9                  & -                     & 70.9                  & 56.1                   \\
DR-CAML                                 & 89.7                  & 98.5          & 8.6                   & 52.9                  & -                     & 69.0                  & 54.8                   \\
MSATT-KG                                & 91.0                  & \textbf{99.2} & 9.0                   & 55.3                  & -                     & 72.8                  & 58.1                   \\
HyperCore                               & 93.0                  & 98.9          & 9.0                   & 55.1                  & -                     & 72.2                  & 57.9                   \\
JointLAAT                               & 92.1                  & 98.8          & 10.7                  & 57.5                  & 80.6                  & 73.5                  & 59.0                   \\ 
\midrule
\textbf{LAAT*}                  & 92.0\tiny{±0.11}                  & 98.8\tiny{±0.02}          & 9.7\tiny{±0.24}                   & \uline{57.4}\tiny{±0.16}          & \uline{81.2}\tiny{±0.22}          & \uline{73.9}\tiny{±0.17}          & 59.0\tiny{±0.14}                   \\
w/ HiCuA          & \textbf{\uline{94.8}}\tiny{±0.07} & \uline{99.1}\tiny{±0.01}  & \uline{10.2}\tiny{±0.21}          & \uline{57.4}\tiny{±0.11}          & \uline{81.2}\tiny{±0.12}          & \uline{73.9}\tiny{±0.10}          & \uline{59.1}\tiny{±0.09}           \\
\midrule
\textbf{RAC*} & 93.0\tiny{±0.08}                  & 98.8\tiny{±0.02}          & 7.9\tiny{±0.30}                   & 55.4\tiny{±0.27}                  & 80.8\tiny{±0.15}                  & 73.2\tiny{±0.18}                  & 57.8\tiny{±0.11}                   \\
w/ HiCuA       & 94.3\tiny{±0.09}                  & \uline{99.0}\tiny{±0.01}  & \uline{8.4}\tiny{±0.17}           & \uline{56.5}\tiny{±0.17}          & \uline{81.2}\tiny{±0.32}          & \uline{73.8}\tiny{±0.17}          & \uline{58.8}\tiny{±0.12}           \\

w/ HiCuC       & \uline{94.4}\tiny{±0.15}          & \uline{99.0}\tiny{±0.01}  & \uline{8.4}\tiny{±0.54}           & 55.8\tiny{±0.44}                  & 81.1\tiny{±0.22}                  & 73.6\tiny{±0.20}                  & 58.6\tiny{±0.12}                   \\
\midrule
\textbf{MultiResCNN*}           & 91.2\tiny{±0.23}                  & 98.7\tiny{±0.02}          & 8.6\tiny{±0.40}                   & 56.2\tiny{±0.34}                  & 81.7\tiny{±0.17}                  & 74.3\tiny{±0.20}                  & 59.1\tiny{±0.22}                   \\
w/ HiCuA      & \uline{94.7}\tiny{±0.10}          & \uline{99.1}\tiny{±0.02}  & 9.2\tiny{±0.33}                   & 56.7\tiny{±0.29}                  & 82.0\tiny{±0.14}                  & 74.8\tiny{±0.16}                  & 59.6\tiny{±0.07}                   \\
w/ HiCuC      & 94.6\tiny{±0.12}                  & \uline{99.1}\tiny{±0.01}  & 9.3\tiny{±0.55}                   & 56.6\tiny{±0.45}                  & 82.1\tiny{±0.11}                  & 74.8\tiny{±0.17}                  & 59.6\tiny{±0.14}                   \\
w/ HiCuA+ASL  & 93.7\tiny{±0.20}                  & 98.9\tiny{±0.02}          & 11.4\tiny{±0.36}                  & \textbf{\uline{57.6}}\tiny{±0.13} & \textbf{\uline{82.4}}\tiny{±0.16} & \textbf{\uline{75.1}}\tiny{±0.14} & \textbf{\uline{59.8}}\tiny{±0.12}  \\
w/ HiCuC+ASL  & 94.0\tiny{±0.33}                  & 98.9\tiny{±0.04}          & \textbf{\uline{11.5}}\tiny{±0.41} & 57.4\tiny{±0.21}                  & \textbf{\uline{82.4}}\tiny{±0.25} & \textbf{\uline{75.1}}\tiny{±0.19} & 59.7\tiny{±0.09}                   \\
\bottomrule
\end{tabular}
\end{table}

\subsection{Results}
Following \citep{li2020icd, mullenbach2018explainable}, we compared the results respectively in full-code and top-50-code MIMIC-III datasets. 
We show both the original and enhanced results for each selected encoder architecture with our HiCu algorithm. ``HiCuA'' indicates our HiCu algorithm with the addition method for the hyperbolic correction part, while ``HiCuC'' indicates the same algorithm with the concatenation method. ASL indicates using the asymmetric loss instead of the standard binary cross-entropy loss. 

\paragraph{HiCu improves generalization across all metrics, over convolution, recurrent and transformer-based neural networks}
The full-code results in Table~\ref{tab:fullcode} show that compared to LAAT, RAC reader and MultiResCNN, we find that our algorithm improves performances across all metrics.\footnote{The re-implementation details of the * models are listed in appendix \ref{sec:implementation}.} 
By leveraging the label hierarchy to generate curricula, HiCu pushes the performance of the \emph{same} encoders enabling them to generalize significantly better than they would without the use of curriculum learning. 
The models trained with HiCu result in the strongest performance among all models on all metrics except Micro-AUC. 
Our overall best model (MultiResCNN w/ HiCuA+ASL) outperforms its corresponding baseline MultiResCNN and improves Macro-AUC by 2.7\%, Micro-AUC by 0.2\%, Macro-F1 by 32.6\%, Micro-F1 by 2.5\%, P@5 by 0.9\%, P@8 by 1.1\%, and P@15 by 1.2\%.

\begin{figure}[]
    \centering
    \includegraphics[width=0.7\textwidth]{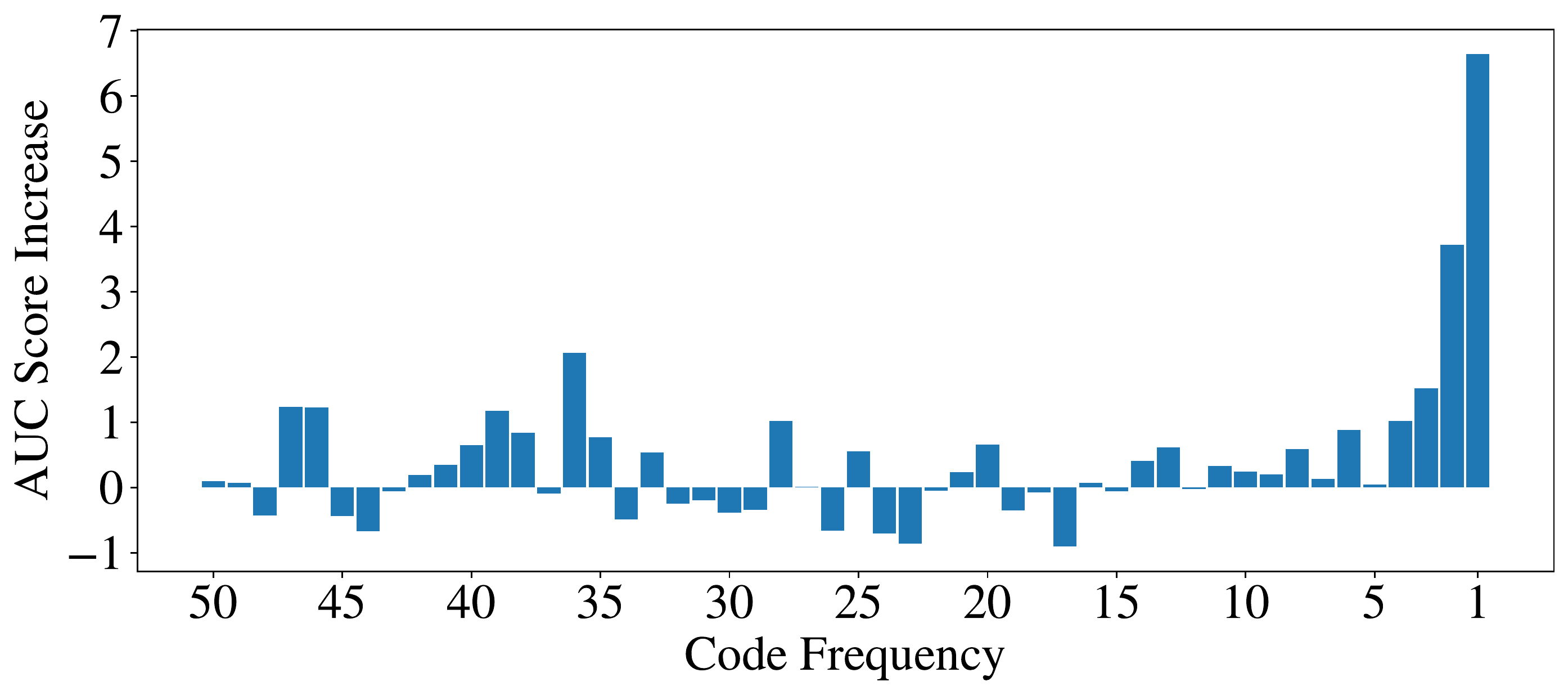}
    \caption{\small \textbf{AUC Improvements on Rare Codes}: AUC Score difference between vanilla MultiResCNN and (MultiResCNN w/ HiCuA+ASL). ICD codes are grouped by their frequency (i.e. occurring times in ground truth).
    This figure shows the increase of average AUC scores (in \%) inside of each frequency group. For example, the size of the bar plot above code frequency "1" indicates the increase of average AUC scores within the code subset that appears only once in the training set. 
    The figure indicates a visible enhancement on rare code prediction. 
    }
    \label{fig:AUC_score}
\end{figure}

\paragraph{HiCu's improvements in performance are most visible among rare labels} In multi-label classification, each label is not equally likely to occur. Indeed the distribution over ICD codes is long-tailed. The results from Table \ref{tab:fullcode} indicate that we observed improvements in the Macro metrics; this suggests that our algorithm can better handle the long-tail of ICD codes and improve predictive performance on rare labels. 

To better understand how our HiCu algorithm improves the Macro AUC/F1 scores, we plot the increase of average AUC scores between the vanilla MultiResCNN model and our (MultiResCNN w/ HiCuA+ASL) model as a function of ICD code frequency in the full-code test dataset (Figure~\ref{fig:AUC_score}). For the rare codes that occur less than or equal to fifty times in the ground truth, we group these codes based on their frequency and compute the average AUC increases. It is worth mentioning that these rare labels already cover 43.1\% of the labels in the dataset, and around 45.7\% of labels never appear in the ground truth of the test dataset. The results show that gains in AUC scores at rarer labels are emphasized. For the codes with a frequency less than 15, we see a coherent AUC increase, indicating that our HiCu algorithm significantly improves classification rates across the rare labels.


\paragraph{HiCu's performance improves with label size}
To quantify the effect of the size of the label set, we study the predictive performance of the 50-code version of the dataset. In this dataset, the numbers of nodes in different tree levels are approximately the same. As shown in Table~\ref{tab:50code} (of the supplementary material), we find that our best model obtains competitive results at each metric relative to the vanilla LAAT, MultiResCNN, and RAC models. 
However, we note that the improvements from using HiCu are much more pronounced on the more realistic, full-code version in Table \ref{tab:fullcode}, suggesting that the data regime in our method works best when there is a large label space (and consequently a deeper and wider hierarchy to take advantage of in building curricula). 

\begin{table}
\small
\caption{\small \textbf{Ablation Study Results (in \%).} The results we obtained are shown in means from 10 random runs. MultiResCNN model are used as the baseline here. (KT) indicates the knowledge transfer process. (HCA) and (HCC) respectively indicate the hyperbolic correction with the addition and concatenation method. Steady improvements on different metrics are shown step by step in the table.}
\label{tab:ablation}
\centering
\begin{tabular}{l|cc|cc|ccc} 
\toprule
\multirow{2}{*}{Model} & \multicolumn{2}{c|}{AUC} & \multicolumn{2}{c|}{F1} & \multicolumn{3}{c}{Precision@K}  \\ 
\cline{2-8}
                       & Macro & Micro            & Macro & Micro           & P@5   & P@8   & P@15             \\ 
\midrule
\textbf{MultiResCNN*}                  & 91.2 & 98.7            & 8.6  & 56.2           & 81.7 & 74.3 & 59.1            \\
w/ KT               & 93.8 & 99.0            & 8.9  & 56.4           & 81.6 & 74.3 & 59.1            \\
w/ KT+HCA           & \textbf{94.7} & \textbf{99.1}            & 9.2  & 56.7           & 82.0 & 74.8 & 59.6            \\
w/ KT+HCC           & {94.6} & \textbf{99.1}            & 9.3  & 56.6           & 82.1 & 74.8 & 59.6            \\
w/ KT+HCA+ASL    & 93.7 & 98.9            & 11.4 & \textbf{57.6}           & \textbf{82.4} & \textbf{75.1} & \textbf{59.8}           \\
w/ KT+HCC+ASL    & 94.0 & 98.9            & \textbf{11.5} & 57.4           & \textbf{82.4} & \textbf{75.1} & 59.7            \\
\bottomrule
\end{tabular}
\end{table}

\paragraph{Disentangling HiCu's improvements in performance}
\label{sec:ablation}
To better understand how each part of HiCu contributes to the improvements in learning, we perform an ablation study using the MultiResCNN \citep{li2020icd} encoder in Table~\ref{tab:ablation}. Compared to the vanilla MultiResCNN model with a randomly initialized query matrix in the decoder, the introduction of our knowledge transfer mechanism (MultiResCNN w/ KT) improves the model performance for both AUC and F1 metrics and remain competitive for the Precision@K metrics. The improvement is pronounced for the Macro-AUC score, indicating HiCu enables models to better predict rare codes.

Our hyperbolic embedding correction mechanism further provides a steady enhancement to the model performance. Both addition (MultiResCNN w/ KT+HCA) and concatenation (MultiResCNN w/ KT+HCC) operations improve the model performance for all evaluation metrics. 

With the asymmetric loss function (ASL), our model is again further improved. ASL steadily strengthens the performances in F1 scores and Precision@K, which achieve the best in our experiments. Due to ASL's ability to deal with imbalanced data, the improvements are particularly critical in the metrics of Macro F1. Though the AUC scores are slightly lowered, the dramatic improvements in higher F1 scores make this a desirable change.

\begin{figure}[h]
    \centering
    \includegraphics[width=0.75\textwidth]{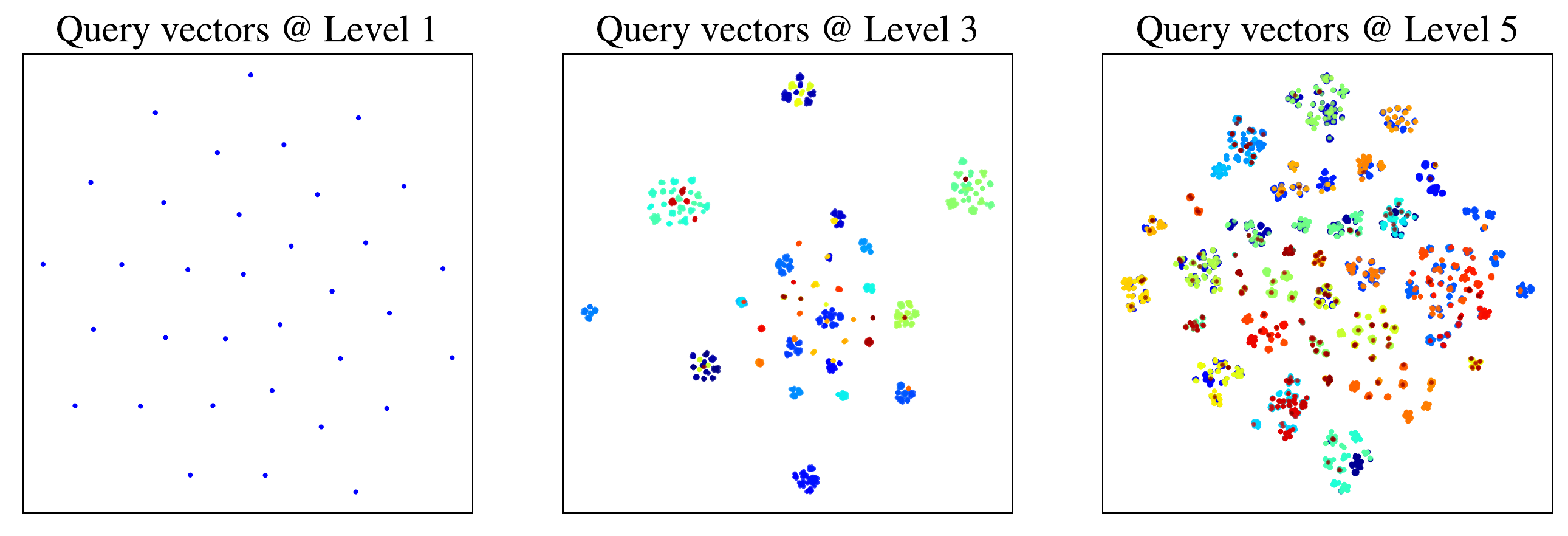}
    \caption{\small Query vectors of the ICD codes at each level in the decoder using the model ``MultiResCNN+HiCuA+ASL''. The query vectors are reduced to two dimensions using t-SNE. Points with the same colors indicate sibling ICD codes under the same parent.}
    \label{fig:tsne}
\end{figure}




\paragraph{Visualizing HiCu}
To interpret the per-label attention mechanism, we visualized the attention matrix $\mathbf{A}$ (as described in Equation~\ref{equ:attention}) given different input discharge summaries, and specifically listed the visualization results for the ICD codes 250.00 (Diabetes mellitus without mention of complication, type II or unspecified type, not stated as uncontrolled) (Figure~\ref{fig:attn1}) and 401.9 (Unspecified essential hypertension) (Figure~\ref{fig:attn2}). Each visualization result is generated by focusing on only one ground true label and one input text, showing the attention weights between the label and the 16 top-weighted tokens in the input. We select several representative results from some of the instances and display them in the appendix. All of the results are based on our selected model (MultiResCNN w/ HiCuA+ASL). The visualization results show that the per-label attention mechanism can consistently focus on relevant words even if the input discharge summaries are different.

To understand how HiCu organizes the representation of the ICD codes during the training process, we visualize the query vectors (see $\mathbf{Q}$ in Equation~\ref{equ:attention}) of the ICD codes at each level of the label tree using t-SNE \citep{van2008visualizing}. As shown in Figure~\ref{fig:tsne}, starting from the first level, the query vectors of the ICD codes are organized in clusters. As more labels are added to downstream levels, sibling codes (codes under the same parent in the label tree) are grouped together. The t-SNE results showcase that the improvements in generalization happen because early steps in the curricula create regions in representation space that are set aside for further child labels to be added. Even if a child label is rare, as long as one or more of its parents is frequent, it will occupy a region in representation space close to similar labels.

%% file: 6_discussions.tex
\section{Discussion} 


HiCu is an easy-to-implement method for curriculum learning that improves predictive performance for multi-label classification across several state-of-the-art choices of encoders. 
For the practical problem of ICD coding, we show that our learning algorithm results in models that generalize better and are more capable of accurately predicting less frequently occurring ICD codes. While there is some theoretical research on curriculum learning \citep{li2022provable, weinshall2020theory}, showing that the gradual increase in depth in the label tree corresponds to an increase in complexity in the learning task at each level would be an interesting direction to investigate the theoretical properties of our algorithm.

\paragraph{Limitations}
One of the key limitations of HiCu is that it requires access to privileged information in the form of structure that is shared among labels. When not readily available, this information can be obtained by querying domain experts. 
In this work, we studied the application of the algorithm to predicting ICD codes -- the codes lie at the leaves of the label tree -- studying the algorithm on multi-label classification problems where the labels lie at intermediate nodes in the hierarchy would be of immediate interest for further investigation. Finally, it is important to acknowledge that while predicting ICD codes, we do not correct for bias in the labels (since ICD codes are created for billing, they may be selected to maximize revenue via reimbursement). 


%% file: appendix.tex
\pagebreak
\appendix
\section*{Appendix}
\addcontentsline{toc}{section}{Appendix}
\renewcommand{\thesubsection}{\Alph{subsection}}

\subsection{Implementation and Hyper-parameter Settings}
\label{sec:implementation}
Our HiCu algorithm was implemented using PyTorch \citep{paszke2019pytorch}.
For the MultiResCNN encoder, we strictly follow the hyper-parameter settings of the encoder introduced in \citet{li2020icd} to train our models. For the Bi-LSTM encoder, We use the official LAAT codebase\footnote{\url{https://github.com/aehrc/LAAT}} instead of our codebase to test the LAAT models as their preprocessing method is different from ours. For the RAC reader encoder, we re-implemented their model since the code has not been publicly released. We follow most of the hyper-parameter settings introduced in \citep{kim2021read} with two modifications. First, we pre-train Word2Vec embeddings for discharge summaries with a minimum word frequency count of 3 instead of 10, as setting this hyper-parameter to 10 results in even worse performance. Second, for all the convolutional filters in the encoder architecture, we set the filter size to 9 instead of 10 to avoid creating an extra dimension for the token features after convolution.

We employ the Adam \citep{kingma2014adam} optimizer for the MultiResCNN and RAC reader models and the AdamW \citep{loshchilov2017decoupled} optimizer for the Bi-LSTM models. Detailed training hyper-parameters can be found in Table~\ref{tab:hyperparam}.
The Bi-LSTM and MultiResCNN-based models are trained using a single NVIDIA Tesla V100 GPU, and the RAC reader-based models are trained using four NVIDIA Tesla V100 GPUs. For ASL loss function hyper-parameters $[\gamma_-, \gamma_+, m]$ (as described in Equation~\ref{equ:asl}), we perform a grid search over different hyper-parameter configurations, ending up with the optimal values [1, 0, 0.05] on the full-code dataset and [1, 0, 0.03] on the 50-code dataset.

\begin{table}[h]
\small
\caption{\small Training hyper-parameters for the full code and top-50 code experiments. ``Input len.'' specifies the input sequence length for each model. ``Emb. dim'' stands for the dimension of Word2Vec embeddings, and ``Tune emb.'' indicates whether to finetune word embeddings during training. ``E.S'' represents early stopping, and ``Sche.'' stands for the training scheduler. ``Epochs'' indicates the different numbers of epochs for the five training rounds.}
\label{tab:hyperparam}
\centering
\begin{tabular}{l|cc|cc|cc} 
\toprule
\multirow{2}{*}{Parameters} & \multicolumn{2}{c|}{Bi-LSTM} & \multicolumn{2}{c|}{MultiResCNN} & \multicolumn{2}{c}{RAC Reader}  \\ 
\cline{2-7}
                            & Full       & 50              & Full        & 50                 & Full       & 50                 \\ 
\midrule
Input len.                    & 4000        & 4000             & 4096         & 4096                & 4096        & 4096                \\
Emb. dim                    & 100        & 100             & 100         & 100                & 300        & 300                \\
Tune emb.                    & No        & No            & Yes         & Yes                & Yes        & Yes               \\
Optimizer                   & AdamW      & AdamW           & Adam        & Adam               & Adam       & Adam               \\
Batch size                  & 8          & 8               & 8          & 8                 & 16         & 16                 \\
Learning rate               & 5e-4       & 5e-4            & 5e-5        & 5e-5               & 8e-5       & 8e-5               \\
Epochs                      & 2,3,3,3,50 & 1,1,1,1,50      & 2,3,5,10,50 & 2,2,3,5,50         & 2,3,5,7,50 & 2,2,3,5,50         \\
E.S. metric                   & P@8        & Micro-F1        & P@8         & P@8                & P@8        & P@8                \\
E.S. patience                 & 6          & 6               & 10          & 10                 & 10         & 10                 \\
Sche. factor                & 0.9        & 0.9             & N/A         & N/A                & N/A        & N/A                \\
Sche. patience              & 2          & 5               & N/A         & N/A                & N/A        & N/A                \\
\bottomrule
\end{tabular}
\end{table}

\pagebreak
\subsection{Supplementary Materials}
\begin{figure}[!h]
  \centering
  \includegraphics[width=1.0\textwidth]{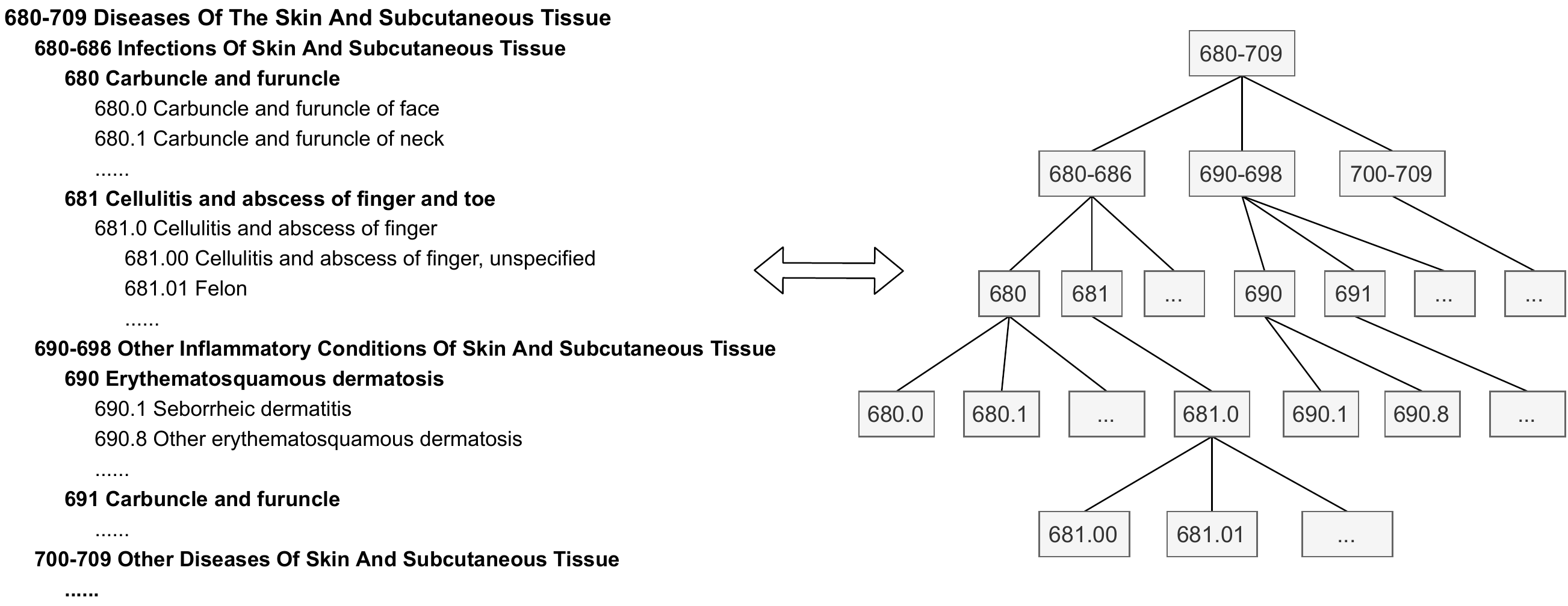}
  \caption{\small An example of representing the ICD-9 taxonomy using a tree structure.}
  \label{fig:icd_taxonomy}
\end{figure}

\begin{algorithm}[!h]
\caption{HiCu: Hierarchical Curriculum Learning}\label{alg:hcl}
\begin{algorithmic}
\STATE $Data=\{\mathbf{x}, \mathbf{y}\}_{i=1}^D \textrm{: training dataset}$
\STATE Enc: text encoder for discharge summary feature extraction
\STATE Dec: decoder network for per-label attention and classification
\STATE $T \textrm{: label tree generated from } \mathbf{y}$
\STATE $\hat{T} \textrm{: augmented label tree generated from } \mathbf{y} \textrm{ and } T$
\STATE $\mathbf{E}_h \textrm{: hyperbolic embeddings for ICD codes generated using } T$
\STATE $T,\ \hat{T} \gets$ GenerateLabelTree($\mathbf{y}$)
\STATE $\mathbf{E}_h \gets$ TrainHyperbolicEmbeddings($T$)

\For{$k$ in range($\hat{T}$.maxLevel + 1)}{
    \For{$i$ in range(nEpochs[$k$])}{
         \For{$\mathbf{x}, \mathbf{y}$ in dataloader($Data$)}{
            \STATE {\qquad $\mathbf{H} \gets$ Enc($\mathbf{x}$)}
            \STATE {\qquad $\mathbf{\bar{y}} \gets$ Dec$_k$($\mathbf{H}$, $\mathbf{E}_h$)}
            \STATE {\qquad Loss $\gets$ ASL($\mathbf{\bar{y}}, \mathbf{y}$)}
            \STATE {\qquad Loss.backward()}
            \STATE {\qquad Optimizer.step()}
            }
    }
    \If{$k \neq \hat{T}$.maxLevel}{
        \STATE Dec$_{k+1}$.$\mathbf{Q}$ $\gets$ KnowledgeTransfer(Dec$_k$.$\mathbf{Q}$)
    }
}
\end{algorithmic}
\end{algorithm}

\begin{table}[!h]
\small
\caption{\small \textbf{MIMIC-III 50 Code Results (in \%).} The results we obtained are shown in means ± standard deviations from 10 random runs. 
The first block in the table shows the baseline results reported in the papers. The following three blocks specify three groups of experiments, each showing the baseline we re-implemented (with *) and the corresponding models we further evaluated with our HiCu learning algorithm. The bold values indicate the best results we obtained from all experiments, whereas the underlined values indicates the best results among each experiment group. 
}
\label{tab:50code}
\centering
\begin{tabular}{l|cc|cc|c} 
\toprule
\multirow{2}{*}{Model} & 
\multicolumn{2}{c|}{AUC} &
\multicolumn{2}{c|}{F1} &
\multicolumn{1}{c}{Precision@K} \\ 
\cline{2-6} & Macro & Micro & Macro & Micro & P@5 \\ 
\midrule
CAML                                                                & 87.5  & 90.9                                                    & 53.2   & 61.4                                                    & 60.9                  \\
DR-CAML                                                             & 88.4  & 91.6                                                    & 57.6   & 63.3                                                    & 61.8                  \\
MSATT-KG                                                            & 91.4  & 93.6                                                    & 63.8   & 68.4                                                    & 64.4                  \\
HyperCore                                                           & 89.5  & 92.9                                                    & 60.9   & 66.3                                                    & 63.2                  \\
JointLAAT                                                           & \textbf{92.5}  & \textbf{94.6}                                                    & 66.1   & \textbf{71.6}                                                    & \textbf{67.1}               \\
\midrule
\textbf{LAAT*} & \underline{92.3}\scriptsize{±0.16}  & \underline{94.4}\scriptsize{±0.11}                                                    & 65.6\scriptsize{±0.45}   & \underline{71.2}\scriptsize{±0.21}                                                    & \underline{67.0}\scriptsize{±0.20}               \\ 
w/ HiCuA+ASL                                                     & 92.1\scriptsize{±0.14}  & 94.2\scriptsize{±0.06}                                                    & \underline{\textbf{66.4}}\scriptsize{±0.37}   & 70.9\scriptsize{±0.26}                                                    & 66.9\scriptsize{±0.12}               \\

\midrule
\textbf{RAC*}                                               & 88.3\scriptsize{±0.06}  & 91.1\scriptsize{±0.06}                                                    & 56.0\scriptsize{±0.31}   & 62.1\scriptsize{±0.33}                                                    & 61.0\scriptsize{±0.14}               \\ 
w/ HiCuA                                                     & \underline{90.8}\scriptsize{±0.09}  & \underline{93.2}\scriptsize{±0.06}                       & \underline{63.1}\scriptsize{±0.50}   & \underline{67.9}\scriptsize{±0.17}                                                   & \underline{64.4}\scriptsize{±0.25}               \\
w/ HiCuC                                                     & 90.7\scriptsize{±0.09}  & 93.0\scriptsize{±0.09}                                                    & 62.0\scriptsize{±1.07}   & 67.2\scriptsize{±0.55}                                                    & 63.9\scriptsize{±0.24}               \\
\midrule
\textbf{MultiResCNN*}                                                         & 89.8\scriptsize{±0.69}  & 92.6\scriptsize{±0.40}                                                    & 60.8\scriptsize{±1.71}   & 66.5\scriptsize{±1.10}                                                    & 63.5\scriptsize{±0.76}               \\ 
w/ HiCuA                                                    & 91.1\scriptsize{±0.18}  & 93.5\scriptsize{±0.08}                                                    & 63.1\scriptsize{±0.62}   & 68.2\scriptsize{±0.38}                                                    & 64.4\scriptsize{±0.24}               \\
w/ HiCuC                                                    & 91.4\scriptsize{±0.26}  & 93.7\scriptsize{±0.12}                                                    & 63.3\scriptsize{±0.67}   & 68.5\scriptsize{±0.31}                                                    & 65.2\scriptsize{±0.16}               \\
w/ HiCuA+ASL                                                & 91.4\scriptsize{±0.10}  & 93.6\scriptsize{±0.08}                                                    & 65.3\scriptsize{±0.25}   & 68.7\scriptsize{±0.18}                                                    & 64.9\scriptsize{±0.25}               \\
w/ HiCuC+ASL                                                & \underline{91.7}\scriptsize{±0.06}  & \underline{93.8}\scriptsize{±0.06}                                                    & \underline{65.6}\scriptsize{±0.32}   & \underline{69.0}\scriptsize{±0.21}                                                    & \underline{65.3}\scriptsize{±0.26}               \\

\bottomrule
\end{tabular}
\end{table}

\pagebreak
\subsection{Attention Visualization Results}

\begin{figure}[h]
\centering
    \begin{subfigure}[]{
        \includegraphics[trim=70 520 320 50, clip, width=0.45\textwidth]{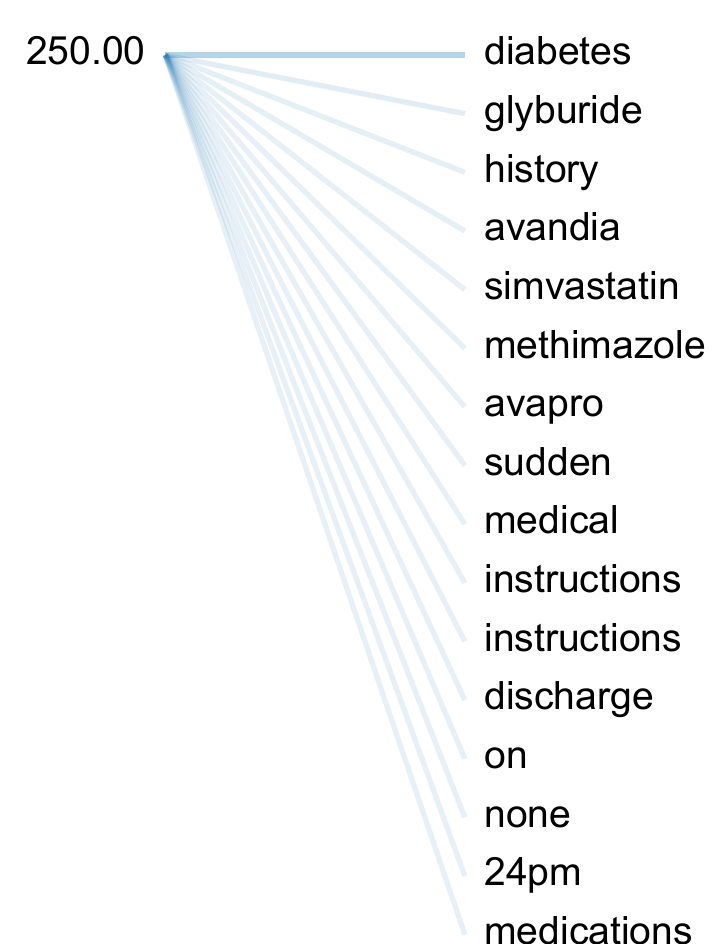}
        \label{fig:attn1_a}}
    \end{subfigure} 
    \hfill
    \begin{subfigure}[]{
        \includegraphics[trim=70 520 320 50, clip, width=0.45\textwidth]{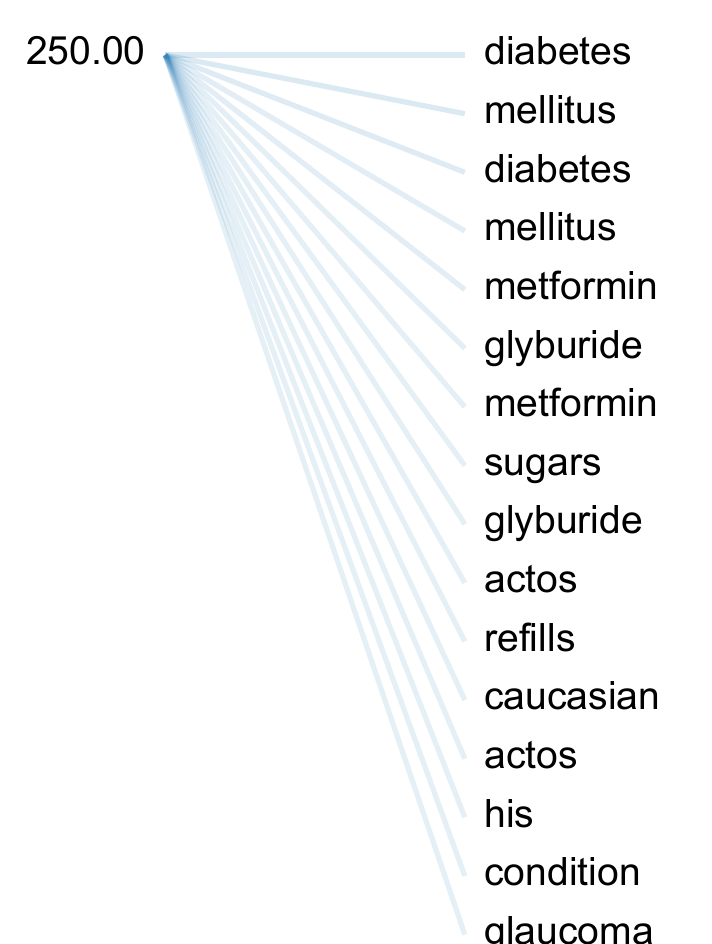}
        \label{fig:attn1_b}}
    \end{subfigure} 
    \hfill
    \caption{\small Attention visualization results for the code 250.00 (type II diabetes) using model ``MultiResCNN+HiCuA+ASL'' trained in the full-code MIMIC-III dataset. The left side is the ground true code, and the right side are 16 top-weighted input tokens ordered from high to low top down. The width of lines suggest the size of the corresponding weight. It is clear that some of the tokens like ``diabetes'', ``glyburide'' coherently occur in the results, which are closely related to the ground true ICD code, and indicate the successful predictions.}
    \label{fig:attn1}
\end{figure}

\begin{figure}[h]
\centering
    \begin{subfigure}[]{
        \includegraphics[trim=70 520 320 50, clip, width=0.45\textwidth]{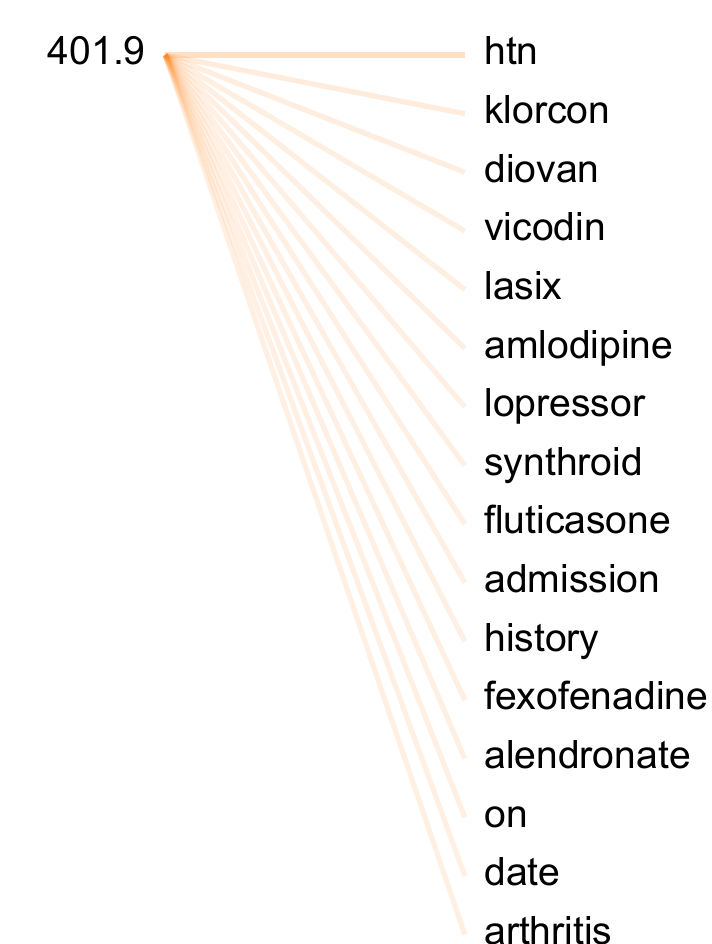}
        \label{fig:attn2_a}}
    \end{subfigure} 
    \hfill
    \begin{subfigure}[]{
        \includegraphics[trim=70 520 320 50, clip, width=0.45\textwidth]{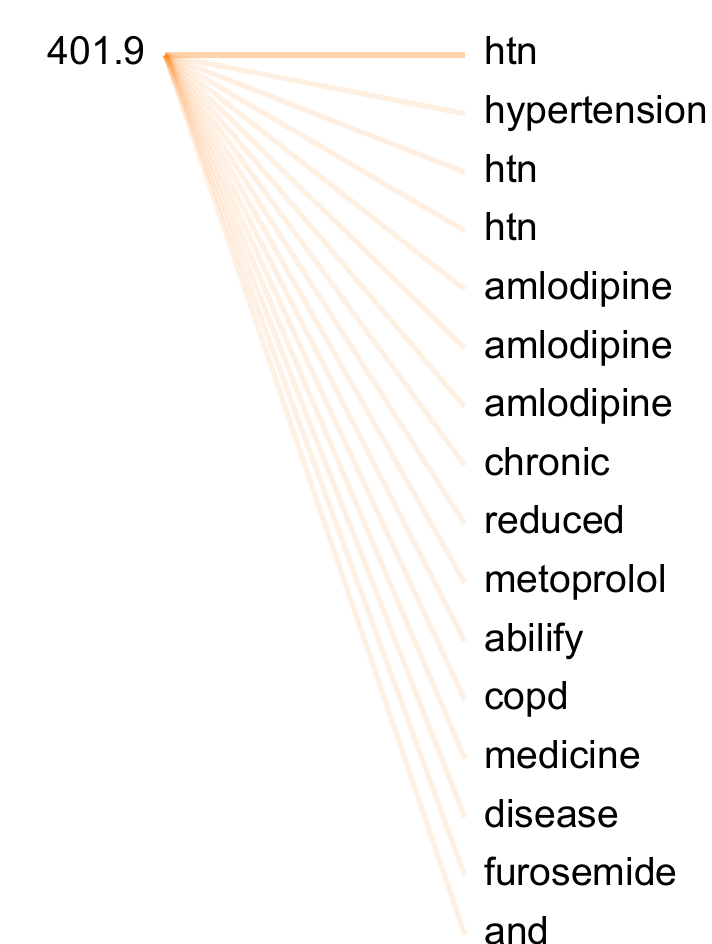}
        \label{fig:attn2_b}}
    \end{subfigure} 
    \hfill
    \caption{\small Attention visualization results for the code 401.9 (unspecified essential hypertension) with model ``MultiResCNN+HiCuA+ASL'' trained in the full-code MIMIC-III dataset. Related tokens like ``htn'' (stands for the abbreviation of hypertension), ``amlodipine'' coherently appear in the results. Most of the top-weighted tokens represent the medicine used for hypertension treatment.}
    \label{fig:attn2}
\end{figure}